# Multiple-image encryption and hiding with an optical diffractive neural network

Yang Gao[1,2], Shuming Jiao[1,2,*], Juncheng Fang[1], Ting Lei[1], Zhenwei Xie[1], Xiaocong Yuan[1,**]
1. Nanophotonics Research Centre, Shenzhen University, Shenzhen, Guangdong, China
2. These authors have equal contributions to this work.
*albertjiaoee@126.com
**xcyuan@szu.edu.cn

*Abstract:*
A cascaded phase-only mask architecture (or an optical diffractive neural network) can be employed for different optical information processing tasks such as pattern recognition, orbital angular momentum (OAM) mode conversion, image salience detection and image encryption. However, for optical encryption and watermarking applications, such a system usually cannot process multiple pairs of input images and output images simultaneously. In our proposed scheme, multiple input images can be simultaneously fed to an optical diffractive neural network (DNN) system and each corresponding output image will be displayed in a non-overlap sub-region in the output imaging plane. Each input image undergoes a different optical transform in an independent channel within the same system. The multiple cascaded phase masks in the system can be effectively optimized by a wavefront matching algorithm. Similar to recent optical pattern recognition and mode conversion works, the orthogonality property is employed to design a multiplexed DNN.

1. Introduction

Information security is a very critical issue in the acquisition, transmission and processing of various types of data such as images. In addition to the commonly used digital security technologies [1] solely based on computer algorithms, the research of optical image encryption, authentication and watermarking techniques [2-5] has received increasingly more attention in recent years. Optical security techniques exhibit certain potential advantages over digital techniques such as high parallelism, high processing speed and direct processing of physical objects.

One common optical architecture for image encryption and watermarking is the cascaded phase-only mask architecture [6-13], which can be dated back to the early work of Double Random Phase Encoding (DRPE) [6] two decades ago. Multiple cascaded phase-only masks (or referred to as phase-only holograms) are aligned at certain separation intervals either in a lens system (such as [6]) or a lensless system (such as [7]). Under coherent light illumination, the diffractive field of the input image is modulated by each phase-only mask sequentially during the forward propagation. Finally, the imaging result in the output plane is recorded as the ciphertext or retrieved hidden image. In many systems (such as [11-13]), the phase-only masks need to be appropriately designed to ensure the system output result is the target one.

In fact, as an optical information processing system, the cascaded phase-only mask architecture has been employed for optical mode conversion [14-18], optical pattern recognition [19,20] and optical image salience detection [21] in previous works, in addition to optical image security. A cascaded phase mask system is also similar to the concept of volume hologram [22] in some extent. In the work [19], a number digit recognition system is implemented all optically with a cascaded phase-only mask architecture, or referred to as an optical diffractive neural network. The phase-only masks in these systems are usually designed by certain optimization algorithms such as Gerchberg–Saxton iteration [11-13], wavefront matching [16] and error back-propagation [19].

In optical security, a cascaded phase mask system can be employed as an encryption system [6-10], which transforms an original image into a noise-like ciphertext. The multiple phase-only masks can be used as the encryption and decryption keys. Unauthorized users without knowing the key cannot recover the original image from the ciphertext. Moreover, a cascaded phase mask system can be employed as an image hiding system [11-13]. The host image is input to the system and another hidden image will be displayed at the output imaging plane, which is a natural image completely different from the input one. Unauthorized users will not able to disclose the hidden image from the host image without a set of correct phase masks.

Conventionally, one such optical security system [6-13] with a given set of phase-only masks can only perform encryption or image hiding for a specific pair of input-output images at one time. To overcome this limitation, the encryption or hiding of multiple images using the same optical architecture has been extensively investigated [23-28]. However, in previous works [23-27], the system usually only performs a transformation from multiple input images to one single output and the output signals from individual input are mixed. In this work, we propose a novel multiplexing system to perform a transformation from multiple input images to multiple output images in a parallel manner, which was seldom investigated previously in the optical security field. To certain extent, this work can be considered as a generalization of the previous work [13]. The previous work [13] was generalized to a multiple-image security system by serially cascading individual systems originally for single input image and single output image [28]. However, the multiple input images in the system [28] will have different priorities in terms of security levels. For example, for two images A and B, A can only be decrypted after B is already decrypted but B can be decrypted when A is not known. As an advantage, the multiple images are independent, multiplexed and have equal security priorities in our proposed system in this work.

In our work, the processing of each different input-output image pair can be considered as an independent "channel", which can be favorable in many practical applications. For example, in some previous works [29-32], optical security systems are designed for biometric authentication such as face recognition and fingerprint verification. The system can display the password image to the user if the user's biometric feature is successfully authenticated. However, one system may be only used for the verification of one specific user. As a significant advantage, our proposed system can support the verification for multiple users and display different output message to each individual user in a multiplexed manner. One possible application scenario of our proposed scheme is shown in Fig. 1.

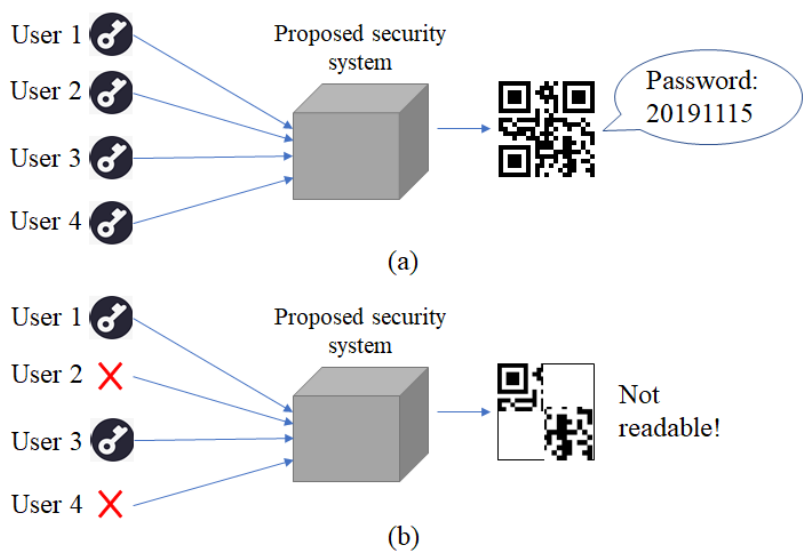

Fig. 1. One possible application scenario of our proposed scheme: (a) the complete QR code image can be reconstructed and the password can be retrieved when all the four users input their correct keys; (b) only some parts of the QR code image can be reconstructed and the password will not be known when only two users input their correct keys.

This paper is organized as follows. The working principles of our proposed system will be described in Section 2. The simulation results and discussions will be given in Section 3. A brief conclusion will be made in Section 4.

2. Proposed Multiple-image Security System with Optical Diffractive Neural Network

2.1 General Framework of Our Proposed System

In our proposed scheme, multiple host images, with each being multiplied by a different phase mask, can be simultaneously input to the system. Under coherent light illumination, the superposed diffractive field will propagate and pass through multiple phase-only masks sequentially. Finally, multiple hidden images will be displayed in non-overlap sub-windows in the output imaging plane. The transformation of each input host image to its corresponding output hidden image is performed in an independent channel with the same set of cascaded phase masks.

A general framework of our proposed system is illustrated in Fig. 2(a). It is assumed that our proposed scheme is designed to simultaneously transform K input host images $I_1, I_2,..., I_{K-1}$ and $I_K$ into K output hidden images $O_1, O_2,..., O_{K-1}$ and $O_K$. In our proposed system, a different random phase-only mask is placed in each input port (at the position of the input imaging plane) and there are totally K different random phase masks. The random phase masks $R_1, R_2,...,R_{K-1}$ and $R_K$ can be arbitrarily generated as long as they are random and different from each other. The reason why each input image is modulated by a different random phase mask first can be explained as follows. After the multiplication with random phase masks, $I_1 R_1$, $I_2 R_2$, …, $I_K R_K$ will become approximately orthogonal to each other, which means the inner product between any two of them (e.g. between $I_1 R_1$ and $I_3 R_3$) will be approximately zero. It shall be noted that the phase masks have to be random and different from each other. If the masks are the same for different images, the orthogonality cannot be achieved. In a cascaded phase-only mask system, orthogonal inputs and orthogonal outputs are favorable for a multiplexed processing, as discussed in [33].

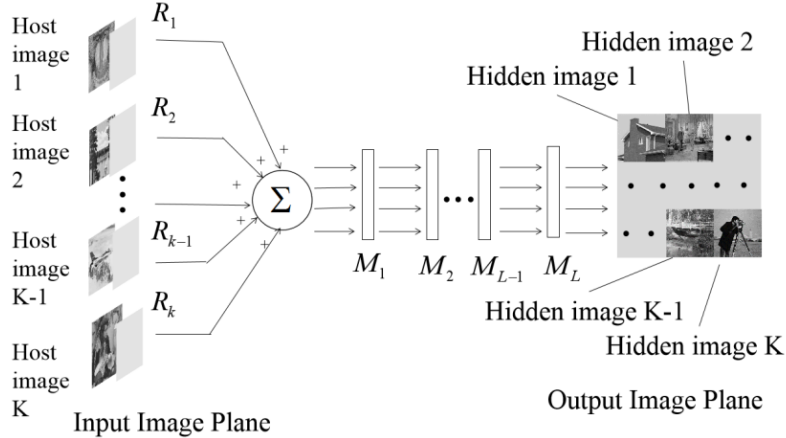

(a)

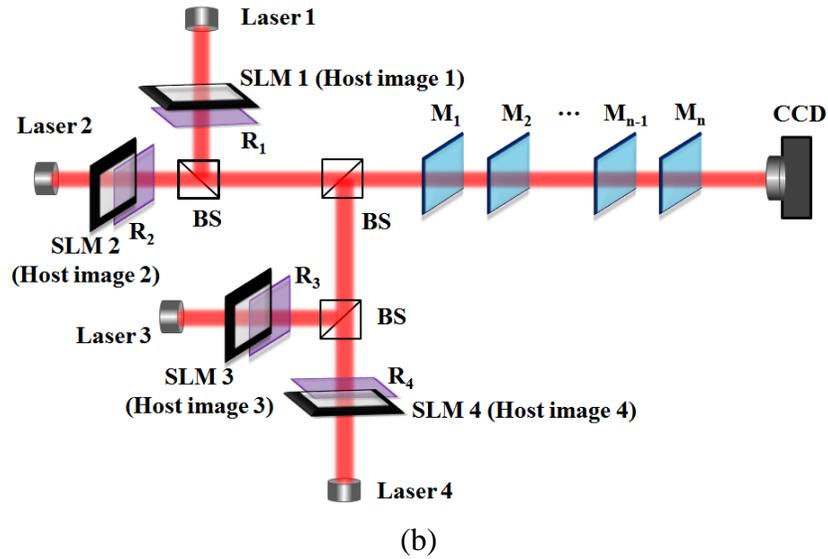

(b)

Fig. 2. (a) Architecture of our proposed multiplexed optical image hiding system (R denotes each random phase mask and M denotes each optimized cascaded phase-only mask); (b) One possible optical setup for practically implementing our proposed scheme (SLM: spatial light modulator; BS: beam splitter; K=4).

Then each input image modulated by its corresponding phase mask is illuminated by a coherent plane wave with the same wavelength. All the diffractive fields will be superposed and propagate from the input imaging plane to the first optimized phase mask $M_1$. The superposition of light fields can be optically implemented with beam combiners, like the superposition of object wave and reference wave in holography. It is assumed the system consists of totally L optimized phase masks ($M_1$, $M_2$, ..., $M_{L-1}$, $M_L$) and for simplicity the neighboring masks are all placed at a distance interval of d. In addition, the distance between the input imaging plane and the mask $M_1$ and the distance between the output imaging plane and the mask $M_L$ are assumed to be d as well. It shall be noticed that the interval distances between masks can be flexibly adjusted and not necessarily all equal. The superposed diffractive field will propagate and pass through each optimized phase mask sequentially. Finally, the light intensity of the diffractive field at the output imaging plane will be recorded. The output imaging plane contains K separate regions and in each region the retrieved hidden image for each corresponding input image will be displayed. The K hidden images are non-overlap and each occupies an identical area. The non-overlapped windows for displaying each output image are orthogonal to each other since they are spatially separate [33].

The most crucial feature of our proposed system is that the processing of each pair of input-output images is independent. The system can accept all the K input host images simultaneously, or only accept one single input host image, or accept an arbitrary subset of all the input host images. The corresponding output images will be displayed in the pre-designed regions of the output imaging plane and the "deactivated" regions will be blank. In fact, the entire system is invertible if the phase masks are conjugated and the light field propagates from right to left in Fig. 2. The input images can also be recovered from the output sub-window images. In this way, a large-size input image can be recovered from a small-size output image in the proposed system.

One possible optical setup for practically implementing our proposed scheme is shown in Fig. 2(b). Each input host image can be fed to the system with a laser source and an amplitude-type spatial light modulator (SLM). The K random phase masks and M optimized phase masks are fabricated as hardcopies and placed in the corresponding positions in the optical setup. The operation of light field superposition in Fig. 2(a) can be realized by several beam splitters in the optical system. The diffractive light fields from host images can be combined pair-wisely with beam splitters and the distance between each SLM and each beam splitter can be appropriately designed to ensure that the diffractive light field for each host image travels at the same distance to the first optimized phase mask. It is equivalent that one diffractive field propagates for a certain distance first and is then superposed to another field with a beam splitter or one diffractive field is superposed to another field with a beam splitter first and then the superposed field propagates for the same distance. As a linear system, $f(I_1 R_1, d) + f(I_2 R_2, d) = f(I_1 R_1 + I_2 R_2, d)$, where $f(..., d)$ denotes a Fresnel light filed propagation for a distance of d. Consequently, the several beam splitters at different positions can jointly perform the superposition operation in Fig. 2(a) in an equivalent way. Finally, the image intensity of the system output is recorded by a CCD sensor.

One critical issue in our proposed scheme is how to design the L optimized phase-only masks to ensure that the system can yield the desired output imaging results. In our work, a wavefront matching algorithm [16] is employed to optimize the phase values of each pixel in the phase-only masks. This algorithm is different from the commonly used Gerchberg–Saxton algorithm [11-13, 23-28] in previous relevant works.

2.2 Wavefront Matching Algorithm for Designing the Optimized Phase-only Masks

The wavefront matching algorithm for designing the optimized phase masks in our proposed system is described as follows. A simple system only consisting of two phase-only masks $M_1$ and $M_2$ is taken as an example for explanation. For simplicity, it is assumed that the system is designed for multiplexing two pairs of host images and hidden images ($I_1$ and $O_1$, $I_2$ and $O_2$, shown in Fig. 3). The number of phase masks and the number of image pairs can be far more than two. But the working principles remain the same.

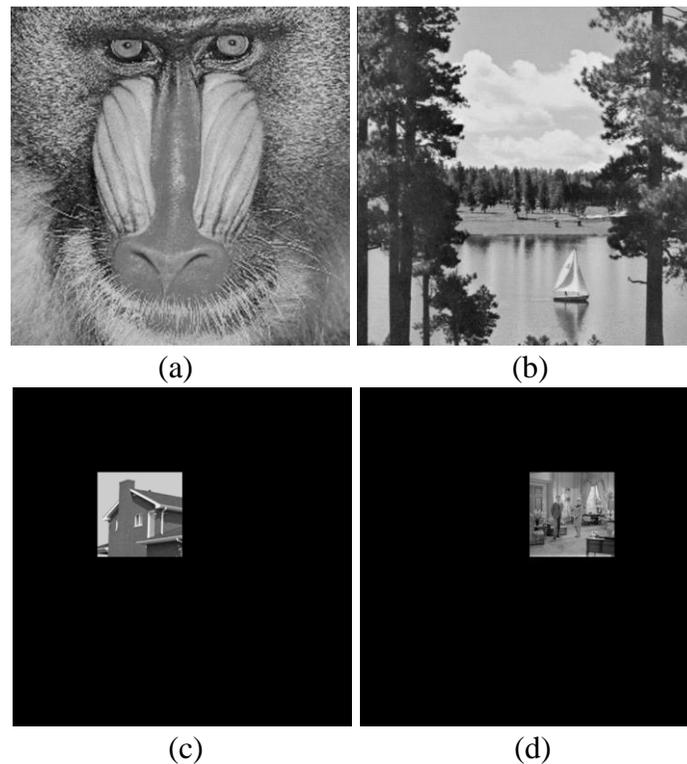

Fig. 3. Two pairs of input images and output images for multiplexed image hiding (a) Input host image$I_1$; (b) Input host image$I_2$; (c) Output hidden image $O_1$ ; (d) Output hidden image $O_2$.

As stated above, the two host images $I_1$ and $I_2$, multiplied with corresponding phase masks $R_1$ and $R_2$, can be simultaneously input to the system as $(I_1 R_1 + I_2 R_2)$ or each one can be individually input to the system, shown in Fig. 4. The intensity of the output field will vary according to the actual input.

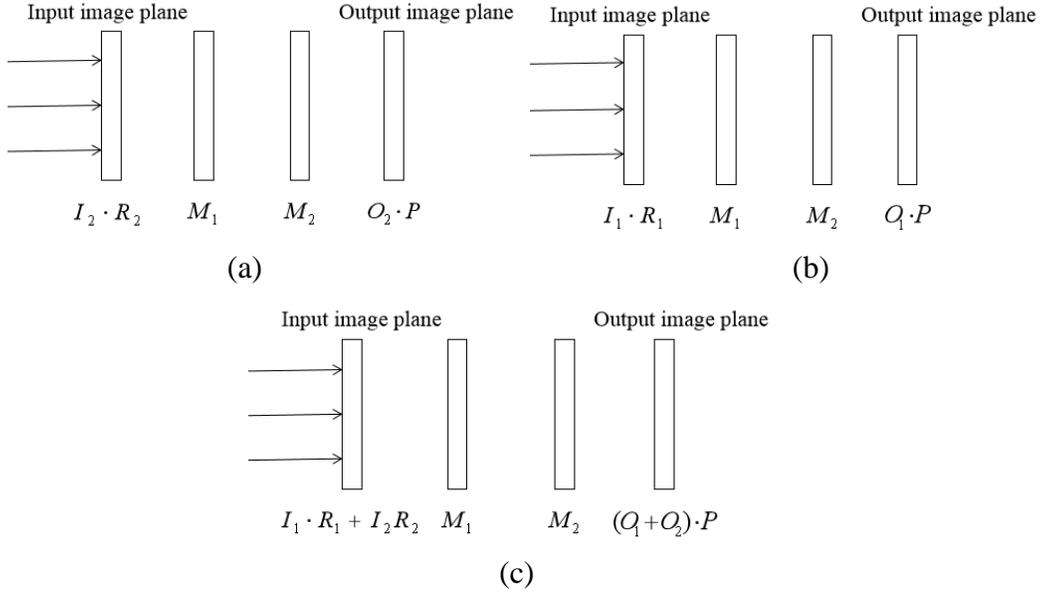

Fig. 4. A simple example of proposed system with only two layers of optimized phase-only masks. The system can multiplex two input-output image pairs and the input can (a) only include image $I_1$ and corresponding random phase mask $R_1$; (b) only include image $I_2$ and corresponding random phase mask $R_2$; (c) include both.

The optimization of phase-only masks is implemented iteratively. In each iteration, there will be three steps (L+1 steps if there are L cascaded phase masks in the system). Initially, $M_1$, $M_2$ and P are set to be random phase masks or uniform phase masks. They will be updated in each iteration whereas $R_1$ and $R_2$ are unchanged. P can be considered as a "virtual phase mask" representing all the phase distributions of different output imaging sub-windows, which is not an actual optimization constraint.

Step One: the input field propagates forward to the back of $M_1$ and the output field propagates backward to the front of $M_1$. Each input host image and output hidden image is processed individually, illustrated in Equations (1)-(4), with f (…, d) representing a free-space Fresnel field propagation by the angular spectrum method with distance d (either forward propagation with a positive value or backward propagation with a negative value). The phase mask $M_1$ is updated according to the rules in Equation (5), with Phase[...] representing the retrieval of phase part and conj(…) representing the phase conjugation. The mathematical proof of Equation (5) can be found in the appendix part.

$$D_{forward1} = f(I_1 R_1, d) \quad (1)$$
$$D_{forward2} = f(I_2 R_2, d) \quad (2)$$
$$D_{back1} = f[f(O_1 P, -d)conj(M_2), -d] \quad (3)$$
$$D_{back2} = f[f(O_2 P, -d)conj(M_2), -d] \quad (4)$$
$$M_1 = Phase[D_{back1}conj(D_{forward1}) + D_{back2}conj(D_{forward2})] \quad (5)$$

Step Two: similar to Step One, the input field propagates forward to the back of $M_2$ and the output field propagates backward to the front of $M_2$. The phase mask $M_2$ is updated according to (6), as:

$$M_2 = Phase\{f(O_1 P, -d)conj[f(f(I_1 R_1, d)M_1, d)] + f(O_2 P, -d)conj[f(f(I_2 R_2, d)M_1, d)]\} \quad (6)$$

Step Three: the input field propagates forward to the output plane and the output phase P is updated according to (7), as:

$$P = Phase[O_1 f(f(f(I_1 R_1, d)M_1, d)M_2, d) + O_2 f(f(f(I_2 R_2, d)M_1, d)M_2, d)] \quad (7)$$

After Step Three, the algorithm will enter the next iteration and starts with Step One again. After a certain number of iterations, the algorithm will finally converge and the optimized phase-only masks $M_1$ and $M_2$ are finally obtained. The wavefront matching algorithm for phase mask optimization is illustrated in Fig. 5.

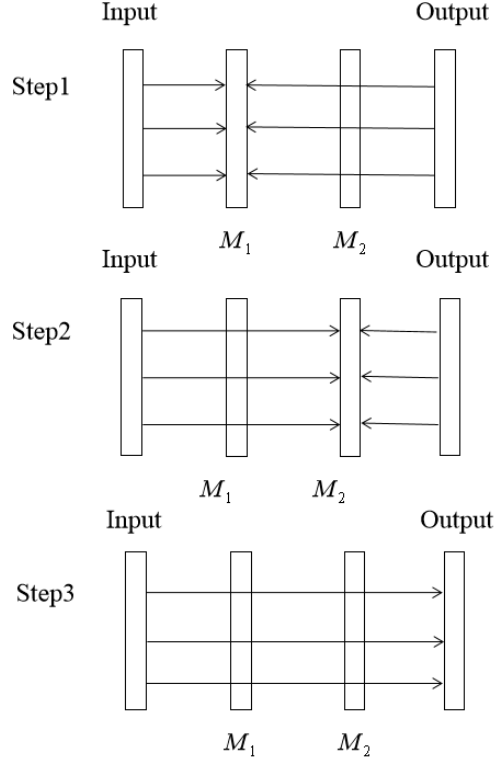

Fig. 5. Wavefront matching algorithm for phase mask optimization in our proposed system

In the simulation of Fresnel field propagation in this work, the angular spectrum method is employed for the generation of all the results. It shall be noted that there are two ways, i.e. circular convolution and linear convolution [34], for the implementation of angular spectrum method. It is assumed that both the input light field before propagation and the output light field after propagation have $N \times N$ pixels. In circular convolution, the input plane is directly transformed to the output plane with an angular spectrum consisting of $N \times N$ pixels. In linear convolution, the input plane is first expanded to a plane containing $2N \times 2N$ pixels by zero-padding. Then it is transformed to the output plane with an angular spectrum consisting of $2N \times 2N$ pixels. Finally, the center window with $N \times N$ pixels is cropped from the output plane as the final Fresnel transformed result. A comparison between circular convolution and linear convolution in the angular spectrum method is shown in Fig. 6. Linear convolution is more accurate than circular convolution in the modeling of light field propagation. But the forward and backward propagation will not be identical due to the cropping operation in the linear convolution, which is not favorable in the iterative optimization of phase masks in the wavefront matching algorithm. In this work, both circular convolution and linear convolution are used. In the iterative design of phase masks, circular-convolution-based angular spectrum method is adopted. After the optimized phase masks are designed, the output result is obtained from the input by modeling the system with linear-convolution-based angular spectrum method. The transfer function in the angular spectrum method can be also filtered in a bandlimited way for more accurately modeling the far-field propagation [34]. Since the propagation distance between neighboring phase masks is small in this work, the transfer function is not modified in the angular spectrum method.

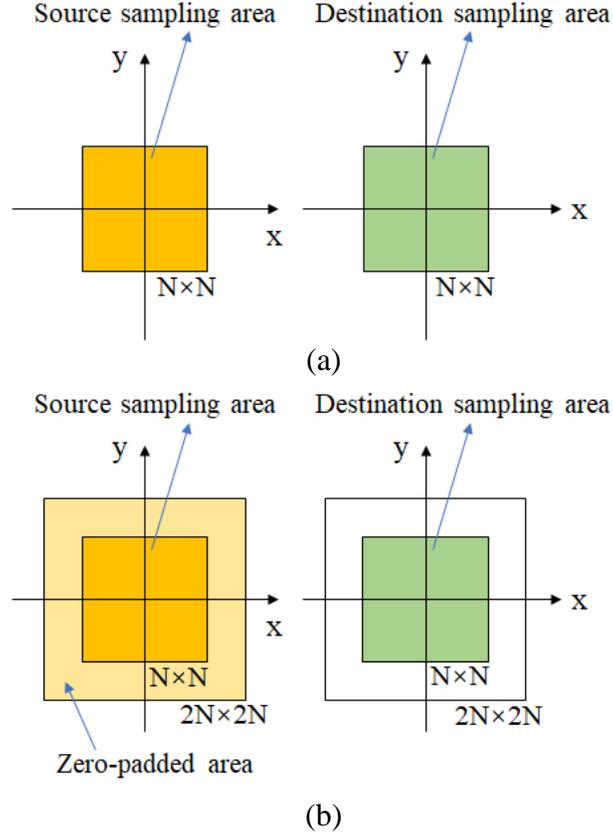

Fig. 6. Comparison of two implementations in the angular spectrum method (a) circular convolution; (b)linear convolution.

In addition, the modulation masks in the system are not necessarily phase-only theoretically. The modulation masks can perform amplitude-only modulation and complex modulation (both amplitude and phase). Equation (5) will be changed to Equation (8) and Equation (9) respectively for amplitude masks and amplitude-phase masks (Real[…] refers to the real part of the light field). Equation (6) needs to be modified accordingly as well. The mathematical proof can be found in the appendix part.

$$M_1 = Real[D_{back1}conj(D_{forward1}) + D_{back2}conj(D_{forward2})]/(D^2_{forward1} + D^2_{forward2}) \qquad (8)$$

$$M_1 = [D_{back1}conj(D_{forward1}) + D_{back2}conj(D_{forward2})]/(D^2_{forward1} + D^2_{forward2}) \qquad (9)$$

3. Result and Discussion

In the simulation, four input-output image pairs are tested with our proposed multiplexed optical image hiding system consisting of four cascaded optimized phase-only masks. The horizontal and vertical extent of each input image (host image) and each phase mask is 512×512 pixels. The horizontal and vertical extent of each output image (hidden image) is 128×128 pixels. Four non-overlap windows will totally occupy an area of 256×256 pixels in the center of the output imaging plane. The pixel size is 20μm and the separation distance d is 0.05m. The wavelength of the illumination light is 532nm. Four pairs of pre-defined input host images and output hidden images are shown in Fig. 7 and Fig. 8 respectively.

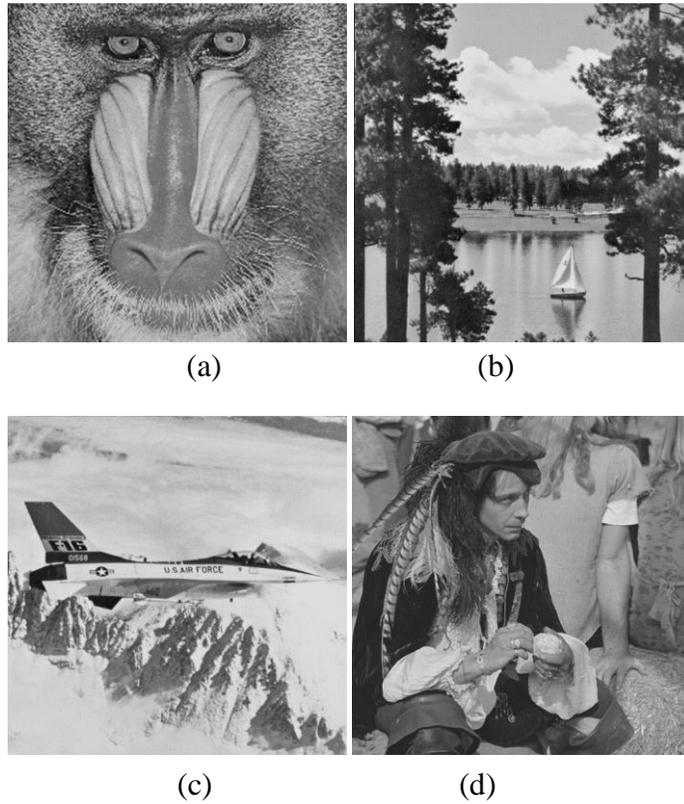

Fig. 7. Four pre-defined input host images in the optical image hiding systems: (a) Input host image $I_1$; (b) Input host image $I_2$; (c) Input host image $I_3$; (d) Input host image $I_4$.

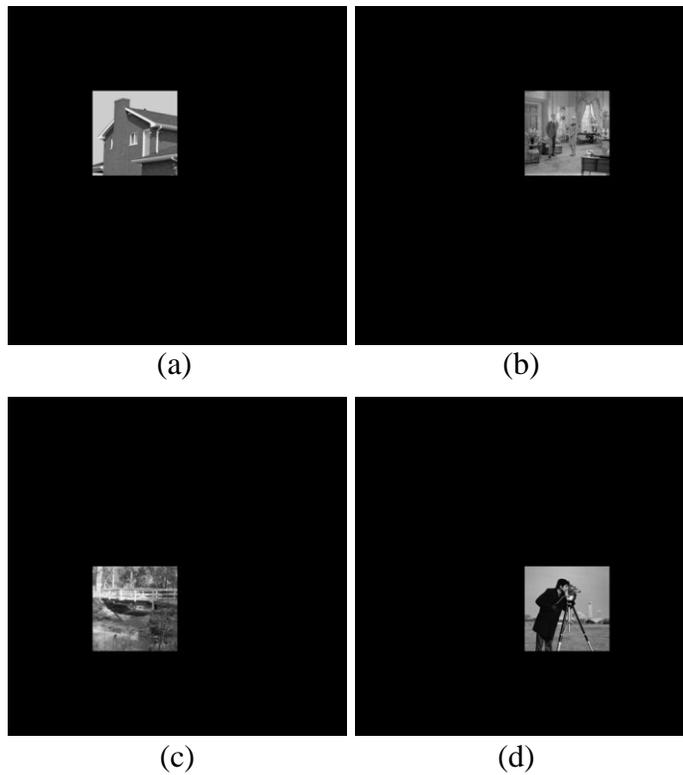

Fig. 8. Four pre-defined output hidden images in the optical image hiding systems (a) Corresponding to Fig. 7(a); (b) Corresponding to Fig. 7(b); (c) Corresponding to Fig. 7(c); (d) Corresponding to Fig. 7(d).

After 30 iterations of optimization using the wavefront matching algorithm, the optimized phase masks can be obtained. The output imaging results are shown in Fig. 9 when the system input is set to be varying host images.

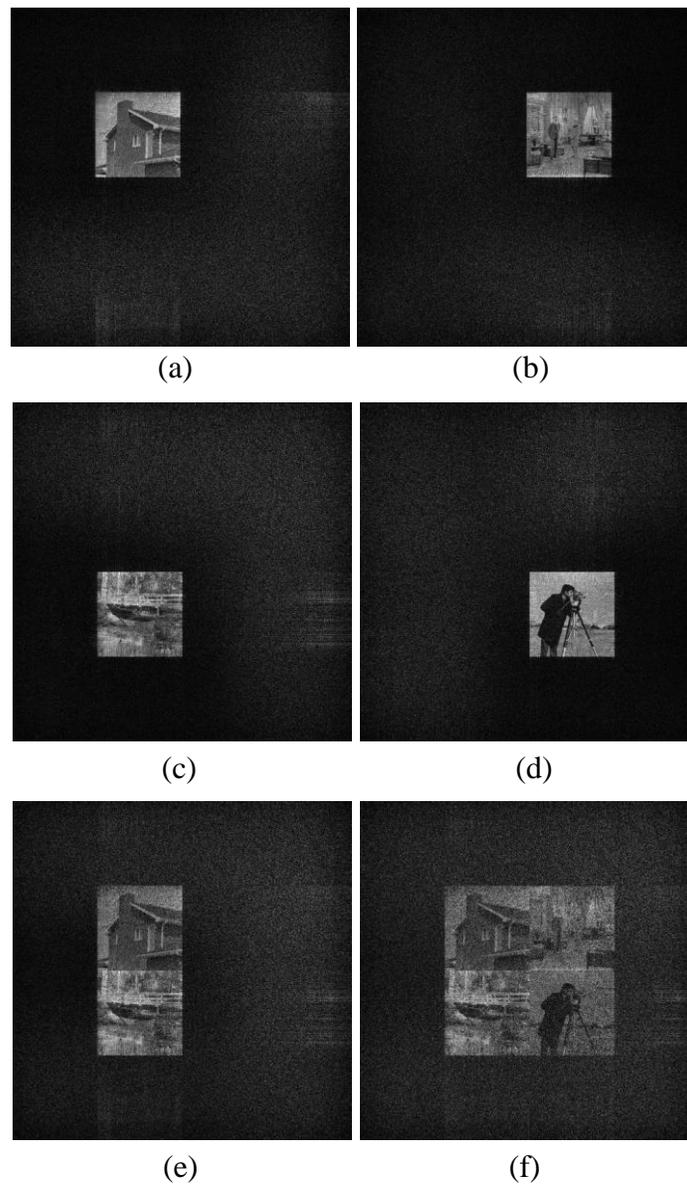

Fig. 9. System output results when the system input is (a) Host image $I_1$; (b) Host image $I_2$;(c) Host image $I_3$;(d) Host image $I_4$;(e) Host image $I_1$ plus Host image $I_3$;(f) All host images.

It can be observed that the designed system can correctly display the desired hidden images in the corresponding regions of the output imaging plane. When a subset of input images is fed into the system, only the corresponding output images will be displayed and the remaining regions are blank. An acceptable level of noise can be observed in the results. The noise problem in optical security systems can be possibly solved by error correction coding [35-36].

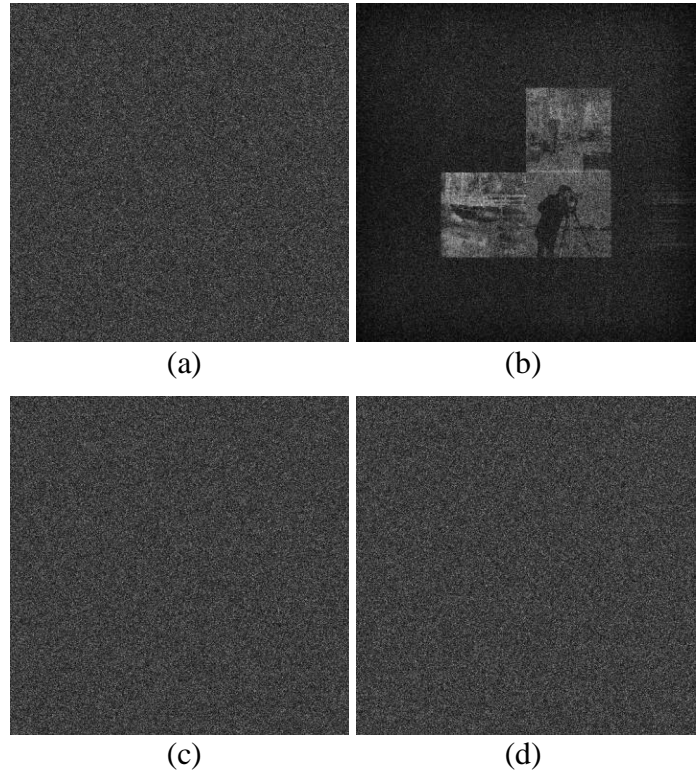

Fig. 10. (a) System output result when the input is host image $I_1$ with wrong $R_1$ mask; (b) System output result when the input includes all the four host images with wrong $R_1$ mask and correct $R_2, R_3$ and $R_4$ masks; (c) System output result when the input is host image $I_1$ with wrong $M_1$ mask; (d) System output result when the input includes all the four host images with wrong $M_1$ mask.

The four random phase masks ($R_1, R_2, R_3$ and $R_4$) and four optimized cascaded phase masks ($M_1, M_2, M_3$ and $M_4$) can be employed as security keys. The desired output images can only be displayed when the input image multiplied with the corresponding random phase mask is correct and all the cascaded phase mask keys are correct. Otherwise, only random-noise results will be obtained in the output imaging plane, shown in Fig. 10.

The number of optimized phase-only masks in the system, i.e. L, is not limited to four and can be reduced or increased. On the one hand, as the number of phase mask layers increases, there will be more degrees of freedom in the light field modulation and transformation to enhance the quality of output imaging results. The mapping function between input images and output images will be simulated more accurately by the optical system. On the other hand, more loss will be introduced due to the limited aperture size of each mask if the light field passes through more phase masks and propagates with a longer distance. The PSNR (Peak Signal to Noise Ratio) of output imaging results compared with the reference images for varying number of phase masks is shown in Fig. 11 (red curve). It can be observed that the quality is optimal when the number of phase masks is L=4. The quality will be degraded either the number of phase masks is less than or more than 4. The output imaging results are shown in Fig. 12(a), Fig. 12(b) and Fig. 12(c) for comparison when the number of phase masks is 2, 4 and 6. The result in Fig. 12(a) is contaminated with more noise compared with Fig. 12(a). The result in Fig. 12(c) is contaminated with less noise compared with Fig. 12(b) but the image information in the area near the four edges in Fig. 12(c) is eroded in some extent. Consequently the overall quality of Fig. 12(c) is worse than Fig. 12(b).

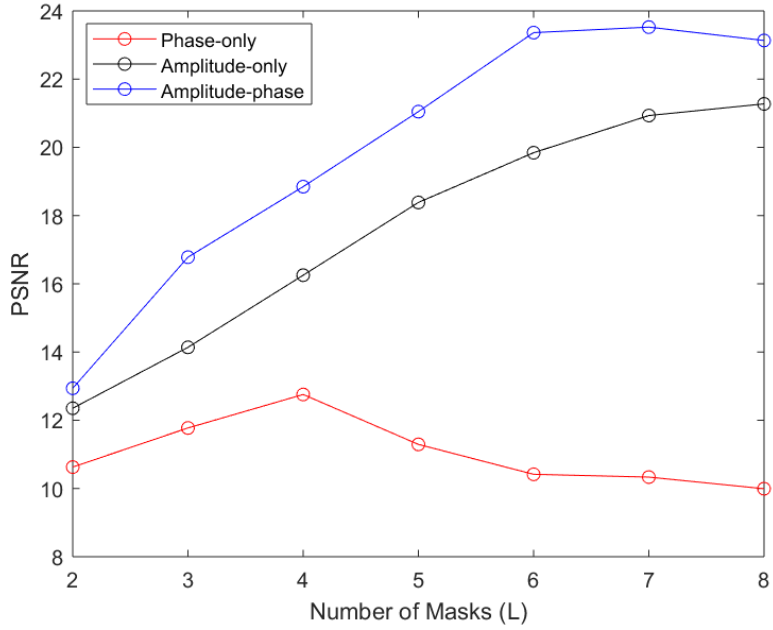

Fig. 11. PSNR (Peak Signal to Noise Ratio) of output imaging results compared with the reference images for varying number of phase-only, amplitude-only and amplitude-phase masks (four input-output channels).

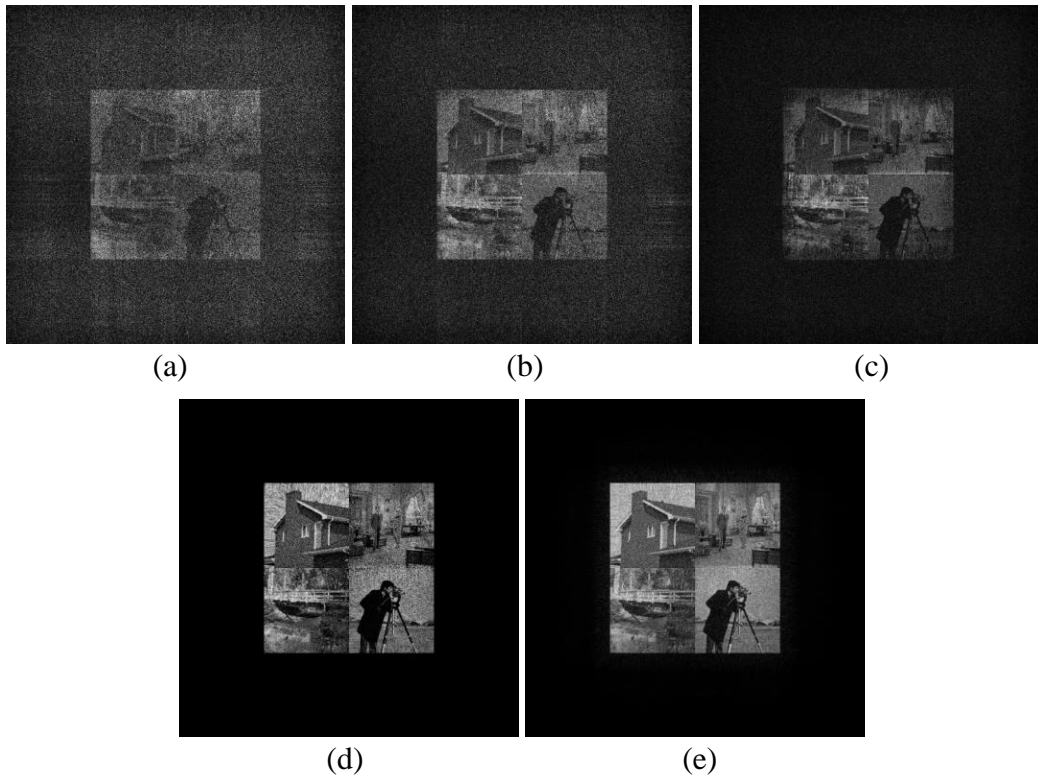

Fig. 12. Output imaging result when all the four host images are input to the system: (a) 2 layer of phase masks (L=2); (b) 4 layers of phase masks (L=4); (c) 6 layers of phase masks (L=6); (d) 4 layers of amplitude masks (L=4); (e) 4 layers of amplitude-phase masks (L=4).

Based on Equation (8), Equation (9) and the appendix part, the proposed system with amplitude-only masks and amplitude-phase masks is simulated as well. From Fig. 11, it can be observed that the amplitude-only masks can yield better results than the phase-only masks and the amplitude-phase masks can yield even better results than the amplitude-only masks. By using the same number of masks (L=4), the output imaging results with amplitude modulation and amplitude-phase modulation are shown in Fig. 12(d) and Fig. 12(e), compared with the one with phase-only modulation in Fig. 12(b). In practice, amplitude-only masks suffer from the problem of low diffraction efficiency and a complex modulation mask for both amplitude and phase is difficult to realize. As a result, phase masks are more commonly used in previous works even though phase-only modulation may not always be the optimal way of modulation.

In addition, the amount of noise in the output imaging results will generally increase when the number of multiplexed input-output image pairs K increases if the number of phase mask layer remains fixed, shown in Fig. 13. The PSNR for Fig. 13(a), Fig. 13(b) and Fig. 13(c) compared with original reference images are 13.7286 dB, 13.3097 dB and 12.7565 dB respectively, as the number of parallelly processed image pairs increase from 2 to 3 and 4. An open source Matlab code for the simulation of our proposed system can be found in [37].

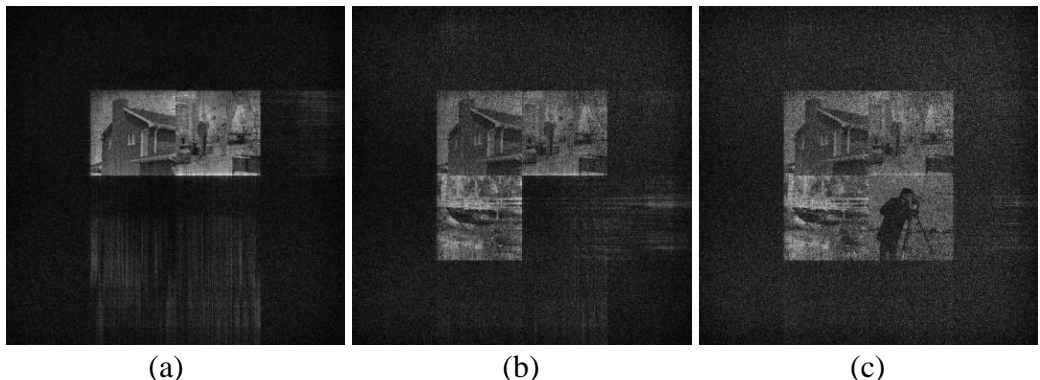

(a)          (b)          (c)

Fig. 13. Output imaging result when the cascaded phase-only mask system contains 4 layers of phase masks (L=4) and K host images are input: (a) K=2; (b) K=3; (c) K=4.

4. Conclusion

An optical architecture with cascaded phase-only masks (or phase-only holograms) is often employed for optical information processing such as optical mode conversion, optical pattern recognition and optical security. The cascaded phase mask system is also referred to as an optical diffractive neural network. In previous works, such an architecture is attempted for transforming one single input image to one single output image, or multiple input images into one single output, for encryption or watermarking. In this work, we propose an optical image hiding (or encryption) system that can process multiple pairs of input-output images in a multiplexed manner for the first time.

Each input host image is multiplied with a random phase mask first and the complex light field can be individually or jointly input to a cascaded phase-only mask architecture with L layers. With the propagating diffractive field being modulated by each phase mask sequentially, the desired output results will be displayed in each spatially separated sub-windows of the output imaging plane.

In this work, a wavefront matching algorithm is employed to optimize the L phase masks in the system, different from the commonly used Gerchberg–Saxton algorithm. The algorithm can be used to design a cascaded phase mask system with arbitrary number of layers. In contrast, in previous optical security works (such as [2-13]), the number of layers is usually restricted to two to four layers.

Simulation results reveal that our proposed system can achieve acceptable imaging quality, low cross-talk between different channels and good security for image hiding. In future works, the multiplexing capacity may be enhanced by wavelength multiplexing and polarization multiplexing [38]. The implementation of an optical diffractive neural network in an incoherent manner will be investigated as well [39].

**Appendix: wavefront matching algorithm (phase-only, amplitude-only and amplitude-phase)**
It is supposed that $d_{f1}$, $d_{f2}$, $d_{b1}$ and $d_{b2}$ are the complex-amplitudes of one corresponding pixel in two pairs of forward and backward propagating fields $D_{forward1}$, $D_{forward2}$, $D_{back1}$ and $D_{back2}$.

(1) If the mask is a phase-only mask, the corresponding pixel in the mask will be $\exp(j\theta)$. The objective is minimize the overall error $E(\theta)$ between the modulated forward propagating fields and backward propagating fields.

$E(\theta) = |d_{f1}\exp(j\theta) - d_{b1}|^2 + |d_{f2}\exp(j\theta) - d_{b2}|^2$

$= |d_{f1}|^2 + |d_{b1}|^2 - d_{f1}conj(O_1)\exp(j\theta) - d_{b1}conj(d_{f1})\exp(-j\theta)$
$+ |d_{f2}|^2 + |d_{b2}|^2 - d_{f2}conj(O_2)\exp(j\theta) - d_{b2}conj(d_{f2})\exp(-j\theta)$

$= |d_{f1}|^2 + |d_{b1}|^2 - 2Re\{d_{b1}conj(d_{f1})\exp(-j\theta)\}$
$+ |d_{f2}|^2 + |d_{b2}|^2 - 2Re\{d_{b2}conj(d_{f2})\exp(-j\theta)\}$

$= |d_{f1}|^2 + |d_{b1}|^2 + |d_{f2}|^2 + |d_{b2}|^2 - 2Re\{d_{b1}conj(d_{f1})\exp(-j\theta) + d_{b2}conj(d_{f2})\exp(-j\theta)\}$

$= |d_{f1}|^2 + |d_{b1}|^2 + |d_{f2}|^2 + |d_{b2}|^2 - 2Re\{[d_{b1}conj(d_{f1}) + d_{b2}conj(d_{f2})]\exp(-j\theta)\}$

When $E(\theta)$ is minimum, $Re\{[d_{b1}conj(d_{f1}) + d_{b2}conj(d_{f2})]\exp(-j\theta)\}$ is maximum and the angle of $[d_{b1}conj(d_{f1}) + d_{b2}conj(d_{f2})]\exp(-j\theta)$ is zero. Therefore $\theta$ equals the angle of $[d_{b1}conj(d_{f1}) + d_{b2}conj(d_{f2})]$ and $\exp(j\theta) = Phase[d_{b1}conj(d_{f1}) + d_{b2}conj(d_{f2})]$

When the number of input-output pairs is extended from 2 to N,

$$\exp(j\theta) = Phase[\sum_{n=1}^{N} d_{bn}conj(d_{fn})]$$

(2) If the mask is an amplitude-only mask, the corresponding pixel in the mask has an intensity Q. The objective is minimize the overall error $E(Q)$ between the modulated forward propagating fields and backward propagating fields.

$E(Q) = |Qd_{f1} - d_{b1}|^2 + |Qd_{f2} - d_{b2}|^2$

$= |Qd_{f1}|^2 + |d_{b1}|^2 - Qd_{f1}conj(d_{b1}) - Qd_{b1}conj(d_{f1})$
$+ |Qd_{f2}|^2 + |d_{b2}|^2 - Qd_{f2}conj(d_{b2}) - Qd_{b2}conj(d_{f2})$

$= Q^2|d_{f1}|^2 + |d_{b1}|^2 - 2QRe[d_{b1}conj(d_{f1})] + Q^2|d_{f2}|^2 + |d_{b2}|^2 - 2QRe[d_{b2}conj(d_{f2})]$

$= Q^2(|d_{f1}|^2 + |d_{f2}|^2) - 2QRe[d_{b1}conj(d_{f1}) + d_{b2}conj(d_{f2})] + |d_{b1}|^2 + |d_{b2}|^2$

The derivative of $E(Q)$ over Q is $E(Q)' = 2Q(|d_{f1}|^2 + |d_{f2}|^2) - 2Re[d_{b1}conj(d_{f1}) + d_{b2}conj(d_{f2})]$

When $E(Q)$ is minimum, $E(Q)' = 0$ and

$$Q = Re[d_{b1}conj(d_{f1}) + d_{b2}conj(d_{f2})]/(|d_{f1}|^2 + |d_{f2}|^2)$$

When the number of input-output pairs is extended from 2 to N,

$$Q = Re\left[\sum_{n=1}^{N} d_{bn}conj(d_{fn})\right] / \sum_{n=1}^{N} |d_{fn}|^2$$

(3) If the mask is an amplitude-phase modulation mask, the corresponding pixel in the mask will be $Q \cdot \exp(j\theta)$. The objective is minimize the overall error $E(Q, \theta)$ between the modulated forward propagating fields and backward propagating fields.

$E(Q, \theta) = |Qd_{f1}\exp(j\theta) - d_{b1}|^2 + |Qd_{f2}\exp(j\theta) - d_{b2}|^2$

$= |Qd_{f1}|^2 + |d_{b1}|^2 - Qd_{f1}conj(d_{b1})\exp(j\theta) - Qd_{b1}conj(d_{f1})\exp(-j\theta)$

$$+|Qd_{f2}|^2+|d_{b2}|^2 - Qd_{f2}conj(d_{b2})\exp(j\theta) - Qd_{b2}conj(d_{f2})\exp(-j\theta)$$
$$=Q^2|d_{f1}|^2+|d_{b1}|^2 - 2QRe[d_{b1}conj(d_{f1})\exp(-j\theta)]$$
$$+Q^2|d_{f2}|^2+|d_{b2}|^2 - 2QRe[d_{b2}conj(d_{f2})\exp(-j\theta)]$$
$$=Q^2\left(|d_{f1}|^2 + |d_{f2}|^2\right) - 2QRe\{[d_{b1}conj(d_{f1}) + d_{b1}conj(d_{f1})]\exp(-j\theta)\}+|d_{b1}|^2+|d_{b2}|^2$$

The variable $\theta$ is considered first. When $E(Q,\theta)$ is minimum, $Re\{[d_{b1}conj(d_{f1}) + d_{b2}conj(d_{f2})]\exp(-j\theta)\}$ is maximum since $Q \geq 0$ and the angle of $[d_{b1}conj(d_{f1}) + d_{b2}conj(d_{f2})]\exp(-j\theta)$ is zero. Therefore $\theta$ equals the angle of $[d_{b1}conj(d_{f1}) + d_{b2}conj(d_{f2})]$ and $\exp(j\theta) = Phase[d_{b1}conj(d_{f1}) + d_{b2}conj(d_{f2})]$
When the number of input-output pairs is extended from 2 to N,
$$\exp(j\theta) = Phase[\sum_{n=1}^{N} d_{bn}conj(d_{fn})]$$
Then $E(Q,\theta) = Q^2\left(|d_{f1}|^2 + |d_{f2}|^2\right) - 2Q|d_{b1}conj(d_{f1}) + d_{b2}conj(d_{f2})|+|d_{b1}|^2+|d_{b2}|^2$
The partial derivative of $E(M,\theta)$ over Q is $\partial E/\partial Q = 2Q\left(|d_{f1}|^2 + |d_{f2}|^2\right) - 2|d_{b1}conj(d_{f1}) + d_{b2}conj(d_{f2})|$
When $E(Q,\theta)$ is minimum, $\partial E/\partial Q = 0$ and
$$Q = |d_{b1}conj(d_{f1}) + d_{b2}conj(d_{f2})|/\left(|d_{f1}|^2 + |d_{f2}|^2\right)$$
When the number of input-output pairs is extended from 2 to N,
$$Q = |\sum_{n=1}^{N} d_{bn}conj(d_{fn})|/\sum_{n=1}^{N}|d_{fn}|^2$$
Then entire complex-amplitude modulation mask can also be written as:
$$Q \cdot \exp(j\theta) = \sum_{n=1}^{N} d_{bn}conj(d_{fn}) / \sum_{n=1}^{N}|d_{fn}|^2$$

Acknowledgement
National Natural Science Foundation of China (11774240, 61805145); Leading Talents Program of Guangdong Province (00201505); Natural Science Foundation of Guangdong Province (2016A030312010); Science, Technology and Innovation Commission of Shenzhen Municipality (KQJSCX20170727100838364)

Reference
[1] B. Furht, E. Muharemagic, and D. Socek, "Multimedia encryption and watermarking," Springer Science and Business Media, 2006.
[2] B. Javidil, A. Carnicer, M. Yamaguchi, T. Nomura, E. Pérez-Cabré, M. S. Millán, N. K. Nishchal, R. Torroba, J. F. Barrera, W. Q. He, X. Peng, A. Stern, Y. Rivenson, A. Alfalou1, C. Brosseau, C. L. Guo, J. T. Sheridan, G. H. Situ1, M. Naruse, T. Matsumoto, I. Juvells, E. Tajahuerce, J. Lancis, W. Chen, X. D. Chen, P. W. H. Pinkse, A. P. Mosk, and A. Markman. " Roadmap on optical security," Journal of Optics, vol. 18, no. 8, pp. 083001, 2016.
[3] W. Chen, B. Javidi, and X. Chen, "Advances in optical security systems," Advances in Optics and Photonics, vol. 6, no. 2, pp. 120-155, 2014.
[4] S. Liu, C. Guo, and J. T. Sheridan, "A review of optical image encryption techniques," Optics & Laser

Technology, no. 57, pp. 327-342, 2014.
[5] S. Jiao, C. Zhou, Y. Shi, W. Zou, and X. Li, "Review on optical image hiding and watermarking techniques," Optics & Laser Technology, no. 109, pp. 370-380, 2019.
[6] P. Refregier, and B. Javidi, "Optical image encryption based on input plane Fourier plane random encoding," Optics Letters, vol. 20, no. 7, pp. 767– 769, 1995.
[7] G. Situ, and J. Zhang, "Double random-phase encoding in the Fresnel domain," Optics Letters, vol.29, no. 14, pp. 1584-1586, 2004.
[8] W. Chen, X. Chen, and C. J. Sheppard, "Optical image encryption based on diffractive imaging," Optics Letters, vol. 35, no. 22, pp.3817-3819, 2010.
[9] Y. Qin, Q. Gong, and Z. Wang, "Simplified optical image encryption approach using single diffraction pattern in diffractive-imaging-based scheme," Optics Express, vol. 22, no. 18, pp. 21790-21799, 2014.
[10] E. Ahouzi, W. Zamrani, N. Azami, A. Lizana, J. Campos, and M. J. Yzuel, "Optical triple random-phase encryption," Optical Engineering, vol. 56, no. 11, pp.113-114, 2017.
[11] Z. Liu, L. Xu, Q. Guo, C. Lin, and S. Liu, "Image watermarking by using phase retrieval algorithm in gyrator transform domain," Optics Communications, vol. 283, no. 24, pp. 4923–4927, 2010.
[12] S. Deng, L. Liu, H. Lang, W. Pan, and D. Zhao, "Hiding an image in cascaded Fresnel digital holograms," Chinese Optics Letters, vol. 4, no. 5, pp. 268–271, 2006.
[13] Y. Shi, G. Situ, and J. Zhang, "Optical image hiding in the Fresnel domain," JOSA A, vol.8, no. 6, pp. 569, 2006.
[14] J.-F. Morizur, L. Nicholls, P. Jian, S. Armstrong , N. Treps, B. Hage, M. Hsu, W. Bowen, J. Janousek, and H-A. Bachor, "Programmable unitary spatial mode manipulation," JOSA A, vol. 27, no. 11, pp. 2524-2531, 2010.
[15] G. Labroille, B. Denolle, P. Jian, P. Genevaux, N. Treps, and J-F. Morizur, "Efficient and mode selective spatial mode multiplexer based on multiplane light conversion," Optics Express, vol. 22, no. 13, pp. 15599-15607, 2014.
[16] N. K. Fontaine, R. Ryf, H. Chen, D. Neilson, and J. Carpenter, "Design of high order mode-multiplexers using multiplane light conversion," in European Conference on Optical Communication (ECOC), Gothenburg, pp. 1-3, 2017.
[17] Nicolas K. Fontaine, Roland Ryf, Haoshuo Chen, David T. Neilson, Kwangwoong Kim & Joel Carpenter, Laguerre Gaussian mode sorter, Nature Communications 10, 1865, 2019.
[18] Q. Zhao, S. Hao, Y. Wang, L. Wang, and C. Xu, "Orbital angular momentum detection based on diffractive deep neural network," Opt. Commun. vol. 443, pp. 245-249, 2019.
[19] X. Lin, Y. Rivenson, N. T. Yardimci, M. Veli, Y. Luo, M. Jarrahi, and A. Ozcan, "All-optical machine learning using diffractive deep neural networks," Science, vol.361, no. 6406, pp. 1004-1008, 2018.
[20] J. X. Li, D. Mengu, Y. Luo, Y. Rivenson, and Aydogan Ozcan, "Class-specific differential detection in diffractive optical neural networks improves inference accuracy," Adv. Photon., Vol. 1, no. 4, pp. 046001, 2019.
[21] T. Yan, J. Wu, T. Zhou, H. Xie, F. Xu, J. Fan, L. Fang, X. Lin, and Q. H. Dai, "Fourier-space Diffractive Deep Neural Network," Physical review letters, vol. 123, no. 2, pp. 023901, 2019.
[22] H. Wang, and R. Piestun, "Dynamic 2D implementation of 3D diffractive optics," Optica, vol. 5, no. 10, pp. 1220-1228, 2018.
[23] Y. Shi, G. Situ, and J. Zhang, "Multiple-image hiding in the Fresnel domain," Optics Letters, vol.32, no. 13, pp.1914-1916, 2007.
[24] A. Alfalou, and A. Mansour, "Double random phase encryption scheme to multiplex and simultaneous encode multiple images," Applied Optics, vol. 48, no. 31, pp. 5933-5947, 2009.
[25] B. Deepan, C. Quan, Y. Wang, and C. J. Tay, "Multiple-image encryption by space multiplexing based on compressive sensing and the double-random phase-encoding technique," Applied Optics, vol.53, no. 20, pp.4539-4547, 2014.
[26] X. W. Li, and I. K. Lee, "Modified computational integral imaging-based double image encryption using fractional Fourier transform," Optics and Lasers in Engineering, no. 66, pp. 112-121, 2015.

[27] Q. Wang, Q. Guo, and L. Lei, "Multiple-image encryption system using cascaded phase mask encoding and a modified Gerchberg–Saxton algorithm in gyrator domain," Optics Communications, no. 320, pp. 12-21, 2014.

[28] Y. Xiao, X. Zhou, S. Yuan, Q. Liu, and Y. Li, "Multiple-image optical encryption: an improved encoding approach," Applied Optics 48(14), 2686-2692, 2009.

[29] Q. Wang, A. Alfalou, and C. Brosseau, "New perspectives in face correlation research: a tutorial," Adv. Opt. Photon, no. 9, pp. 1-78, 2017.

[30] H. Suzuki, M. Yamaguchi, M. Yachida, N. Ohyama, H. Tashima, and T. Obi, "Experimental evaluation of fingerprint verification system based on double random phase encoding," Optics Express, vol. 14, no. 5, pp.1755-1766, 2006.

[31] S. Yuan, T. Zhang, X. Zhou, X. Liu, and M. Liu, "An optical authentication system based on encryption technique and multimodal biometrics," Optics & Laser Technology, no. 54, pp. 120-127, 2013.

[32] L.C. Ferri, A. Mayerhoefer, M. Frank, C. Vielhauer, and R. Steinmetz, "Biometric Authentication for ID Cards with Hologram Watermarks," Security and Watermarking of Multimedia Contents IV, vol. 4675, pp. 629–640, SPIE, 2002.

[33] S. Zheng, X. Zeng, S. Xu, and D. Fan, "Orthogonality of Diffractive Deep Neural Networks," arXiv preprint arXiv:1811.03370, 2018.

[34] K. Matsushima, and T. Shimobaba, "Band-limited angular spectrum method for numerical simulation of free-space propagation in far and near fields," Opt. Express 17(22), 19662-19673, 2009.

[35] S. Jiao, W. Zou, and X. Li, "Noise removal for optical holographic encryption from telecommunication engineering perspective," OSA Digital Holography and Three-Dimensional Imaging 2017, pp. Th2A-5, Optical Society of America, 2017.

[36] S. Jiao, Z. Jin, C. Zhou, W. Zou, and X. Li, "Is QR code an optimal data container in optical encryption systems from an error-correction coding perspective?" JOSA A, vol. 35, no. 1, pp. A23-A29, 2017.

[37] https://github.com/szgy66/code/tree/master

[38] J. F. Barrera, R. Henao, M. Tebaldi, R. Torroba, and N. Bolognini, "Multiplexing encrypted data by using polarized light," Opt. Commun. 260, 109–112, 2006.

[39] S. Jiao, J. Feng, Y. Gao, T. Lei, Z. Xie, and X. Yuan, "Optical machine learning with incoherent light and a single-pixel detector," Opt. Lett. 44 (21), 5186-5189, 2019